\documentclass[journal]{IEEEtran}
\usepackage{amsmath}
\usepackage{amsthm}
\usepackage{amssymb}
\usepackage{algorithm}
\usepackage{array}
\usepackage{algpseudocode}
 
\usepackage{booktabs}
\usepackage{caption}
\usepackage{cite}
\usepackage{enumitem}
\usepackage{makecell}   
\usepackage{graphicx}
\usepackage{hyperref} 
\usepackage{multirow}
\usepackage{rotating} 
\usepackage{ragged2e} % for '\RaggedRight' macro
\usepackage{stfloats}
\usepackage{soul}
\usepackage{subcaption}
\usepackage{tabularx}
\usepackage{textcomp}
\usepackage{times}
\usepackage{url}
\usepackage{verbatim}
\usepackage{xcolor}

\newtheorem{theorem}{Theorem}

\newtheorem{definition}{Definition}
\newtheorem{assumption}[theorem]{Assumption}

\newlist{tabitemize}{itemize}{1} % create a bespoke 1-level itemize-like env.
\setlist[tabitemize]{label=\textbullet, nosep, leftmargin=*, 
                     before={\begin{minipage}[t]{\linewidth}\RaggedRight},
						after={\end{minipage}}}

\begin{document}

\title{LiveVal: Time-aware Data Valuation via Adaptive Reference Points}

\author{Jie Xu, Zihan Wu, Cong Wang~\IEEEmembership{Fellow,~IEEE,} and Xiaohua Jia~\IEEEmembership{Fellow,~IEEE}
        % <-this % stops a space
%\thanks{This paper was produced by the City University in Hong Kong. They are in Piscataway, NJ.}% <-this % stops a space
%\thanks{© 2024 IEEE.  Personal use of this material is permitted.  Permission from IEEE must be obtained for all other uses, in any current or future media, including reprinting/republishing this material for advertising or promotional purposes, creating new collective works, for resale or redistribution to servers or lists, or reuse of any copyrighted component of this work in other works.}
}

\markboth{Journal of \LaTeX\ Class Files,~Vol.~14, No.~8, August~2021}%
{Shell \MakeLowercase{\textit{et al.}}: A Sample Article Using IEEEtran.cls for IEEE Journals}


\maketitle

\begin{abstract}
	Time-aware data valuation enhances training efficiency and model robustness, as early detection of harmful samples could prevent months of wasted computation. However, existing methods rely on model retraining or convergence assumptions or fail to capture long-term training dynamics.
	We propose LiveVal, an efficient time-aware data valuation method with three key designs: 
	1) seamless integration with SGD training for efficient data contribution monitoring; 2) reference-based valuation with normalization for reliable benchmark establishment; and 3) adaptive reference point selection for real-time updating with optimized memory usage. 
	We establish theoretical guarantees for LiveVal's stability and prove that its valuations are bounded and directionally aligned with optimization progress. Extensive experiments demonstrate that LiveVal provides efficient data valuation across different modalities and model scales, achieving 180× speedup over traditional methods while maintaining robust detection performance.
\end{abstract}

\section{Introduction}
The advancement of machine learning systems depends on high-quality training data. Data valuation, which quantifies individual data points' contributions to model performance \cite{ghorbani2019data}, enables enhanced training efficiency~\cite{luo2024fast}, improved model robustness \cite{wang2023data}, and collaborative data sharing~\cite{song2019profit,sim2020collaborative}. 

The ability to evaluate data quality during training is crucial for modern large-scale machine learning systems, as early detection of harmful training samples could prevent months of wasted computation on large models. 
However, existing data valuation methods require model convergence or complete training processes, making them unsuitable for streaming data processing and untargeted poisoning attacks where models may never converge. Specifically, Shapley value-based and Leave-One-Out methods ~\cite{jia2018efficient,jia2019towards,kwon2022beta,sun2023shapleyfl} require complete model retraining, which is computationally prohibitive and provides unreliable valuations due to retraining randomness  ~\cite{wu2022davinz,yoon2020data}. Influence function-based methods ~\cite{koh2017understanding} rely on assumptions about loss convexity and model optimality that rarely hold in deep learning ~\cite{hammoudeh2024training}. Local update methods ~\cite{pruthi2020estimating,paul2021deep}, while avoiding complete retraining, only consider immediate gradient information and thus fail to capture long-term training dynamics. 

It is challenging to develop efficient time-aware data valuation. First, tracking time-varying contributions is computationally intensive, requiring $O(NT)$ calculations for $N$ samples over $T$ training steps for each model update. Second, establishing appropriate benchmarks for time-aware contributions is non-trivial, as data points impact different aspects of training (convergence speed, generalization, robustness). Third, natural decaying gradient magnitudes during training lead to unfair valuations, causing later samples to receive disproportionately lower values regardless of their actual importance.

To address these challenges, we propose LiveVal, a time-aware data valuation method with three key innovations. First, to achieve computational efficiency, LiveVal seamlessly integrates with Stochastic Gradient Descent (SGD) training to monitor data contributions during model updates. Second, to establish reliable benchmarks for measuring data quality, LiveVal introduces a reference-based valuation mechanism where reference points serve as comparison baselines, coupled with normalization that compensates for gradient magnitude decay to enable fair comparison of data contributions across different training stages.
Third, to maintain efficient real-time valuation, LiveVal employs an adaptive reference point selection strategy that dynamically updates reference points based on training dynamics, eliminating the need to store all intermediate states while preserving valuation accuracy.

The main contributions of this paper are:
\begin{itemize}
	\item We propose LiveVal, which seamlessly integrates with SGD training to achieve computational efficiency in monitoring data contributions during model updates.
	\item We design an adaptive reference-based valuation mechanism with normalization that enables fair and time-aware data valuation across different training stages.
	\item We establish theoretical guarantees for LiveVal's stability and demonstrate its significant advantages through extensive experiments, achieving 180× speedup over baseline methods.
\end{itemize}

\section{Preliminaries and Problem Definition}
\subsection{Preliminaries}
We formalize the data valuation problem within a standard supervised learning framework. Let $\mathcal{D} = \{(\mathbf{x}_i, y_i)\}_{i=1}^N$ denote the training dataset drawn from distribution $\mathcal{P}$, where $\mathbf{x}_i \in \mathcal{X}$ and $y_i \in \mathcal{Y}$ are input features and labels respectively. $N$ is the total number of training samples. The function $f$, parameterized by $\boldsymbol{\theta} \in \Theta \subseteq \mathbb{R}^d$ maps inputs to predicted outputs.
Given a loss function $L$, the learning objective is to find optimal parameters $\boldsymbol{\theta}^*$ that minimize the expected loss over the underlying data distribution $\mathcal{P}$:
\begin{equation}
	\boldsymbol{\theta}^* = \arg\min_{\boldsymbol{\theta} \in \Theta} \mathbb{E}_{(\mathbf{x},y) \sim \mathcal{P}}[L(f(\mathbf{x};{\boldsymbol{\theta}}), y)].
\end{equation}

\begin{definition}[Mini-batch SGD]
	At each iteration $t$, a mini-batch $\mathcal{B}_t \subset \{1, \ldots, N\}$, is randomly sampled from the training set. The model parameters are then updated as follows:
	\begin{equation}
		\boldsymbol{\theta}_{t} = \boldsymbol{\theta}_{t-1} - \eta_t \frac{1}{\|\mathcal{B}_t\|} \sum_{i \in \mathcal{B}_t} \nabla_{\boldsymbol{\theta}} L(f(\mathbf{x}_i;{\boldsymbol{\theta}_{t-1}}), y_i), 
	\end{equation}
	where $ \eta_t $ is the learning rate at iteration $t$, and $ \nabla_{\boldsymbol{\theta}} L(f(\mathbf{x}_i;{\boldsymbol{\theta}_{t-1}}), y_i) $ represents the gradient of the loss with the parameters $ \boldsymbol{\theta}_{t-1} $ for the data point $ (\mathbf{x}_i, y_i) $. 
\end{definition}

\begin{definition}[Parameter Trajectory]
	The parameter trajectory $\{\boldsymbol{\theta}_t\}_{t=0}^T$ represents the sequence of model parameters during training, where $\boldsymbol{\theta}_0$ is the initial parameter and $\boldsymbol{\theta}_T$ is the final parameter after $T$ training steps.
\end{definition}

\begin{definition}[Influence Function~\cite{koh2017understanding} ]
	The influence of a training point $\mathbf{z}_i=(\mathbf{x}_i,y_i)$ on the loss at a test point $\mathbf{z}_\text{test}$ is:
	\begin{equation}
		\mathcal{I}(\mathbf{z}_\text{test}, \mathbf{x}_i) = -\nabla_{\boldsymbol{\theta}} L(\mathbf{z}_\text{test}, \boldsymbol{\theta}^*)^T H_{\boldsymbol{\theta}^*}^{-1} \nabla_{\boldsymbol{\theta}} L(\mathbf{x}_i, \boldsymbol{\theta}^*),
	\end{equation}	
	where $ H_{\boldsymbol{\theta}^*} =\frac{1}{N} \sum_{i=1}^N \nabla_{\boldsymbol{\theta}}^2 L(\mathbf{x}_i, y_i; \boldsymbol{\theta}^*) $ is the Hessian matrix of the loss function, representing the second-order derivatives of the loss with the parameters $ \boldsymbol{\theta}^* $.
\end{definition}

\subsection{Problem Definition}
Time-aware data valuation requires measuring how data contributions evolve throughout the training process. 
Formally, let $\mathcal{V}_t: \mathcal{D} \times \{\Theta_k\}_{k=1}^t \rightarrow \mathbb{R}$ be a valuation function that maps a data point and model parameter trajectory to a real value at training step $t$. \begin{definition}[Step-wise Data Value]
	The step-wise value of data point $i$ is
	\begin{equation}
		v_i^t = \begin{cases}
			\mathcal{V}_t((\mathbf{x}_i, y_i), \boldsymbol{\theta}_{t-1}, \boldsymbol{\theta}_{\text{ref}}^t)& \text{if } i \in \mathcal{B}_t \\
			0 & \text{otherwise},
		\end{cases}
	\end{equation}
	where $\boldsymbol{\theta}_{\text{ref}}^t$ is the reference point at step $t$.
\end{definition}

\begin{definition}[Cumulative Data Value]
	The cumulative value of data point $i$ up to step $T$ captures the aggregated impact throughout training as:
	\begin{equation}
		v_i^{[0,T]} = \sum_{t=1}^T v_i^t.
	\end{equation}
\end{definition}

\section{Basic Method with a Static Reference Point}\label{secbasic}
Traditional data valuation methods rely on loss computations, which face several limitations: 1) they are sensitive to the choice of loss function and optimization landscape, 2) they require model convergence or retraining to get reliable estimates, and 3) they cannot effectively capture a data point's impact on the overall optimization trajectory. To address these challenges, we propose to directly measure data contributions in parameter space.

Our key insight is that the final model parameters represent the desired destination of the optimization process, and measuring how each gradient update reduces the distance to this target state provides a more direct and robust way to quantify data contributions. This parameter-space perspective enables continuous monitoring of contributions during training without requiring model convergence or retraining and naturally captures how each data point guides optimization. 

Based on this insight, we first propose a basic method using the final model parameters as a static reference point, which is extended to LiveVal with adaptive reference points in Section \ref{secada}.

Our method consists of 1) monitoring a standard SGD training procedure and 2) quantifying each data point's contribution based on its parameter-space trajectory.

\subsection{Model Training} \label{sectrainphase}
The training phase follows the standard SGD optimization procedure while maintaining essential information for subsequent valuation. At each iteration $t$, the model parameters are updated according to:
\begin{equation}\label{eqsgd}
	\boldsymbol{\theta}_{t} = \boldsymbol{\theta}_{t-1} - \eta_t \frac{1}{\|\mathcal{B}_t\|} \sum_{i \in \mathcal{B}_t} \nabla_{\boldsymbol{\theta}} L(f(\mathbf{x}_i;{\boldsymbol{\theta}_{t-1}}), y_i).
\end{equation}
To enable subsequent data valuation, key intermediate parameters are stored, including $\boldsymbol{\theta}_{t-1}$, $\eta_t$, and $\mathcal{B}_t$. Algorithm \ref{alg:liveval_train} describes the training phase of the basic method.

\begin{algorithm}[t]
	\caption{Training Process of LiveVal}
	\label{alg:liveval_train}
	\begin{algorithmic}[1]  
		\Require{Training data $\mathcal{D} = \{(\mathbf{x}_i, y_i)\}_{i=1}^N$, learning rate schedule $\{\eta_t\}_{t=1}^{T}$, batch size $\|\mathcal{B}_t\|$}
		\Ensure{Stored information $A$}
		\State Initialize model parameter $\boldsymbol{\theta}_0$
		\State Initialize storage $A \leftarrow \emptyset$
		\For{$t = 1$ to $T$}
		\State Sample mini-batch $\mathcal{B}_t$ of size $\|\mathcal{B}_t\|$
		\State $\mathbf{g}_t \leftarrow \frac{1}{\|\mathcal{B}_t\|} \sum_{i \in \mathcal{B}_t} \nabla_{\boldsymbol{\theta}} L(f(\mathbf{x}_i;{\boldsymbol{\theta}_{t-1}}), y_i)$
		\State Store $A[t] \leftarrow (\boldsymbol{\theta}_{t-1}, \eta_t, \mathcal{B}_t)$
		\State $\boldsymbol{\theta}_{t} \leftarrow \boldsymbol{\theta}_{t-1} - \eta_t \mathbf{g}_t$
		\EndFor
		\State \Return $A$
	\end{algorithmic}
\end{algorithm}

\subsection{Basic Valuation}
The valuation phase quantifies individual data contributions by measuring how effectively each gradient update moves the model toward the optimal state.

For each data point $i$ in mini-batch $\mathcal{B}_t$, we first compute the optimal direction toward the final model parameter:
\begin{equation}
	\Delta\boldsymbol{\theta}_t = \boldsymbol{\theta}^* - \boldsymbol{\theta}_{t-1},
\end{equation}
which represents the ideal optimization path.

\begin{figure}[t]
	\centering
	\includegraphics[width=0.99\linewidth]{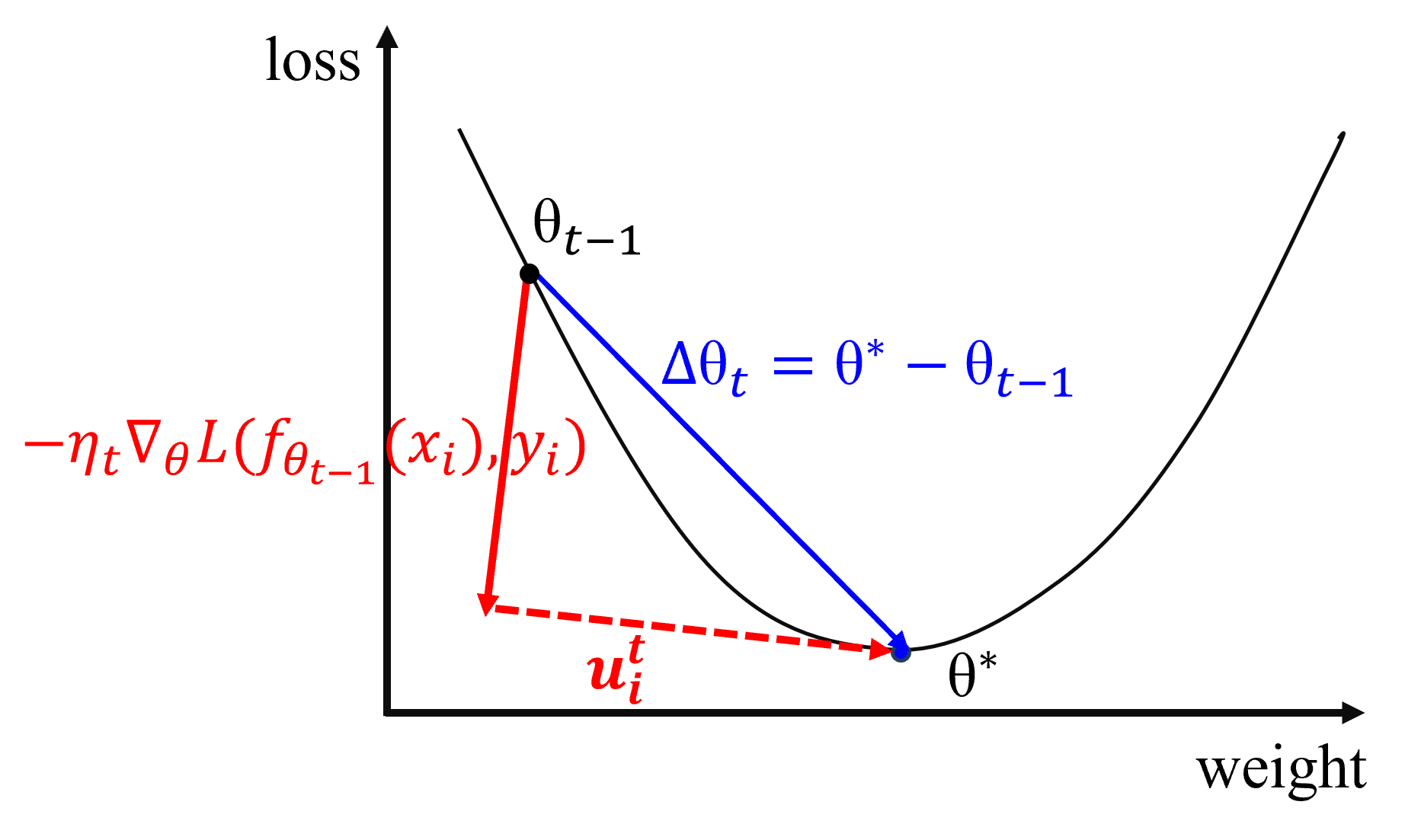}
	\caption{Basic method using the final model parameter as the static reference point.}
	\label{fig:figbasic}
\end{figure}

We then evaluate the data point's contribution through a hypothetical state $\mathbf{u}_i^t$ that measures the remaining distance to $\boldsymbol{\theta}^*$ after applying only data point $i$'s gradient: 
\begin{equation}
	\mathbf{u}_i^t = \boldsymbol{\theta}^* - (\boldsymbol{\theta}_{t-1} - \eta_t \nabla_{\boldsymbol{\theta}} L(f(\mathbf{x}_i;{\boldsymbol{\theta}_{t-1}}), y_i)).
\end{equation}
As shown in Figure \ref{fig:figbasic}, $\|\mathbf{u}_i^t\|$ captures how effectively the gradient moves toward $\boldsymbol{\theta}^*$. A smaller $\|\mathbf{u}_i^t\|$ indicates better alignment with the optimal direction, while larger values suggest the update deviates from the desired trajectory.

The step-wise value is then computed with normalization to ensure fair comparison across training stages:
\begin{equation} \label{eqvit}
	v_i^{t} = \frac{\|\Delta\boldsymbol{\theta}_t\| - \|\mathbf{u}_i^t\|}{\|\Delta\boldsymbol{\theta}_t\|+\|\mathbf{u}_i^t\|}.
\end{equation}
The numerator $\|\Delta\boldsymbol{\theta}_t\| - \|u_i^t\|$ measures improvement toward the reference point. A positive value indicates the data point helps guide the model toward the reference point, while a negative value suggests it pushes the model away.
The denominator $\|\Delta\boldsymbol{\theta}_t\|+\|\mathbf{u}_i^t\|$ normalizes values to ensure fair comparison across training stages. Without this normalization, data points used later in training would be unfairly penalized due to naturally decreasing gradient magnitudes. The detailed analysis is in Section \ref{secprop}.

Finally, we compute the cumulative value:
\begin{equation} \label{eqvitb}
	v_i^{[0,T]} =\sum_{t=1}^T v_i^{t},
\end{equation}
where $v_i^{t}=0$ if $i \notin \mathcal{B}_t$. This aggregation captures the total impact of a data point for the entire training process. The basic time-aware data valuation Algorithm \ref{alg:liveval_infer_until_t} is detailed in Appendix \ref{app:implementation}. 

\section{LiveVal with Adaptive Reference Points} \label{secada}
\subsection{Overview and Motivation}
While effective, the basic method has two limitations: high storage overhead from storing all intermediate parameters and the inability to provide real-time valuations. 
LiveVal addresses these challenges through an adaptive reference point mechanism that 1) enables real-time valuation by using near-future states as reference points, 2) reduces memory requirements by maintaining only a sliding window of parameters, and 3) adapts evaluation horizons based on training dynamics.

\begin{figure}[t]
	\centering
	\includegraphics[width=0.99\linewidth]{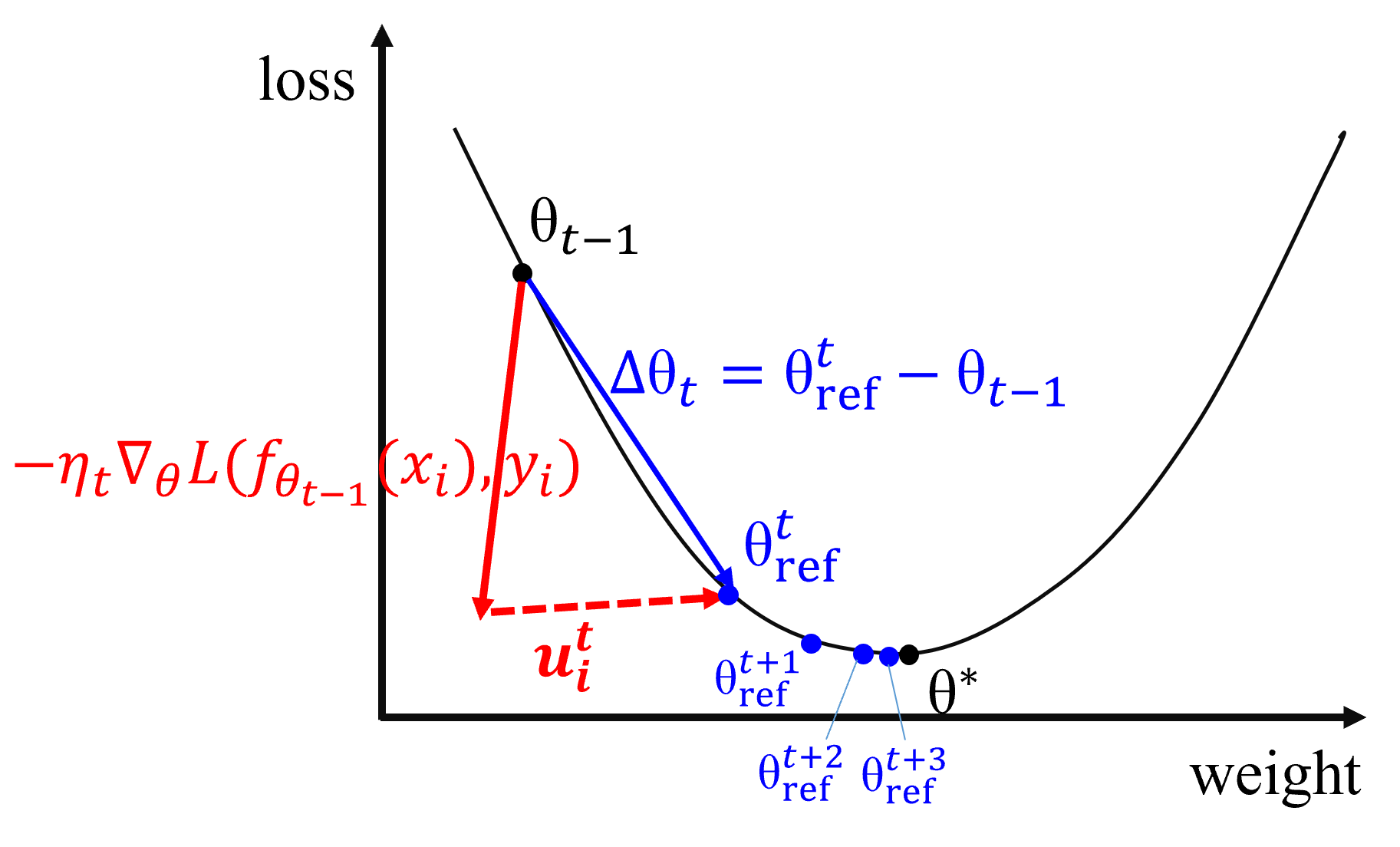}
	\caption{LiveVal's adaptive reference point mechanism. }
	\label{fig:figliveval}
\end{figure}

As shown in Figure \ref{fig:figliveval}, LiveVal dynamically selects reference points within an adaptive window that expands during rapid learning phases and shrinks near convergence.

\subsection{Adaptive Reference Point Mechanism}
The mechanism consists of two key components: dynamic window adjustment and reference point selection.

\subsubsection{Dynamic Window Adjustment}
We define a look-ahead window $\delta_t$ that determines the temporal horizon for the reference point at each step $t$. 

The size of the look-ahead window \( \delta_t \) adapts based on the relative rate of loss change. At each step $t$, we compute:
\begin{equation}
	\dot{L} = \frac{L_{t'} - L_{t-1}}{\delta_{t-1}},
\end{equation}
where $L_{t-1}$ is the loss at step $t-1$,
\begin{equation}
	L_{t-1} = \frac{1}{\|\mathcal{B}_{t-1}\|} \sum_{i \in \mathcal{B}_{t-1}} L(f(\mathbf{x}_i;{\boldsymbol{\theta}_{t-1}}), y_i),
\end{equation}
$t' = \min(t-1+ \delta_{t - 1}, T)$ denotes the step of reference point at step $t-1$, and 
\begin{equation}
	L_{t'} = \frac{1}{\|\mathcal{B}_{t'}\|} \sum_{i \in \mathcal{B}_{t'}} L(f(\mathbf{x}_i;{\boldsymbol{\theta}_{t'}}), y_i).
\end{equation}

The window size adapts according to the loss change rate:
\begin{equation} 
	\delta_{t} = 
	\begin{cases} \min(\delta_{t-1} + \Delta\delta, \delta_{\text{max}}) & \text{if } \|\dot{L}\| > \varepsilon_{\text{max}}, \\
		\max(\delta_{t-1} - \Delta\delta, \delta_{\text{min}}) & \text{if } \|\dot{L}\| < \varepsilon_{\text{min}}, \\
		\delta_{t-1} & \text{otherwise},
	\end{cases} 
\end{equation} 
where $\delta_\text{min}$ and $\delta_\text{max}$ bound the look-ahead window size, $\Delta\delta$ controls the adaptation rate, and $\varepsilon_\text{min}$, $\varepsilon_\text{max}$ are thresholds that trigger adjustments in response to rapid or slow loss changes respectively.

The window size is adjusted according to:
\begin{itemize}
	\item $\|\dot{L} \| > \varepsilon_\text{max}$: During rapid loss decrease, larger windows capture long-term optimization trajectories and avoid misleading measurements from short-term parameter fluctuations.
	\item $\|\dot{L}\| < \varepsilon_\text{min}$: During slow loss decrease, may near convergence, smaller windows identify subtle but valuable optimization effects.
	\item $\varepsilon_\text{min} \leq \|\dot{L} \| \leq \varepsilon_\text{max}$: During steady progress, the window size balances immediate and long-term contributions.
\end{itemize}

\subsubsection{Adaptive Reference Point Selection}
At each step $t$, LiveVal dynamically selects a reference point $\boldsymbol{\theta}_\text{ref}^t$ based on the current window size:
\begin{equation}
	\boldsymbol{\theta}_{\text{ref}}^t = \begin{cases}
		\boldsymbol{\theta}_{t-1+\delta_t} & \text{if } t-1 + \delta_t < T, \\
		\boldsymbol{\theta}_T & \text{otherwise},
	\end{cases}
\end{equation}
where $T$ is the total number of training steps and $\delta_t$ is the adaptive window size. 

This dynamic selection enables real-time valuation while maintaining memory efficiency by only storing parameters within the active window.

\subsection{LiveVal Implementation}
LiveVal achieves real-time data valuation through a novel dual-queue architecture that maintains bounded memory overhead while enabling continuous value computation. The implementation integrates time-aware valuation principles with dynamically adapted reference points for efficient real-time processing.

The core of LiveVal's implementation is a dual-queue system that coordinates real-time valuation with model training. The Model Queue ($Q_{\theta}$) maintains model parameters, batch information, learning rates, and loss values within a sliding window $[t-\delta_{\text{max}}, t]$ where $t$ is the current training step. Rather than storing the complete trajectory requiring $O(Td)$ memory, this design reduces storage to only the actively needed parameters and associated training information. The Reference Queue ($Q_{\text{ref}}$) complements this by tracking $(t_{\text{eval}}, t_{\text{ref}})$ pairs that map evaluation time points to their corresponding reference states.

Value computation in LiveVal extends the basic method through dynamic reference points. At each step, it replaces the static reference point $\boldsymbol{\theta}^*$ with adaptively selected $\boldsymbol{\theta}_\text{ref}^t$ to compute:
\begin{align}
	\Delta \boldsymbol{\theta}_t &= \boldsymbol{\theta}_\text{ref}^t - \boldsymbol{\theta}_{t-1} \\
	\mathbf{u}_i^t &= \boldsymbol{\theta}_\text{ref}^t - (\boldsymbol{\theta}_{t-1} - \eta_t \nabla_{\boldsymbol{\theta}} L(f(\mathbf{x}_i; \boldsymbol{\theta}_{t-1}), y_i)) \\
	v_i^t &= \frac{\|\Delta \boldsymbol{\theta}_t\| - \|\mathbf{u}_i^t\|}{\|\Delta \boldsymbol{\theta}_t\| + \|\mathbf{u}_i^t\|}, \quad
	v_i^{[0,T]} = \sum_{t=1}^T v_i^t
\end{align}
where $v_i^t = 0$ when data point $i$ is not in the current mini-batch $\mathcal{B}_t$.  

\begin{algorithm}[t]
	\caption{LiveVal: Time-aware Data Valuation}
	\label{alg:liveval_integrated}
	\begin{algorithmic}[1]
		\Require Training dataset $\{(\mathbf{x}_i, y_i)\}_{i=1}^N$, Total steps $T$, Initial window $\delta_0$, Window bounds $\delta_\text{min}$, $\delta_\text{max}$, Adaptation rate $\Delta \delta$, Loss thresholds $\varepsilon_\text{min}$, $\varepsilon_\text{max}$
		\Ensure Step-wise values $\{v_i^{t}\}_{t=1}^T$, Cumulative value $v_i^{[0,T]}$
		
		\State Initialize $v_i^{[0,T]} \leftarrow 0$ for all $i$
		\State Initialize array $\{v_i^{t}\}_{t=1}^T \leftarrow \mathbf{0}$ for all $i$
		\State Initialize window size $\delta_0$
		\State Initialize window size array $\{\delta_t\}_{t=1}^T \leftarrow \{\delta_0\}$
		\State Initialize model queue $Q_{\theta}$ for storing model parameters
		\State Initialize reference queue $Q_\text{ref}$ for storing $(t_\text{eval}, t_\text{ref})$ pairs
		\State Initialize $\boldsymbol{\theta}_0$
		
		\For{$t = 1$ to $T$}
		\State Sample mini-batch $\mathcal{B}_{t}$
		\State $\mathbf{g}_{t} \leftarrow \frac{1}{\|\mathcal{B}_{t}\|} \sum_{i \in \mathcal{B}_{t}} \nabla_{\boldsymbol{\theta}} L(f(\mathbf{x}_i;\boldsymbol{\theta}_{t-1}), y_i)$
		\State $\boldsymbol{\theta}_{t} \leftarrow \boldsymbol{\theta}_{t-1} - \eta_{t} \mathbf{g}_{t}$
		\State Store $(t, \boldsymbol{\theta}_{t}, \mathcal{B}_{t}, \eta_{t}, L_{t})$ in $Q_{\theta}$
		\State $t_\text{ref} \leftarrow \min(t-1 + \delta_{t-1}, T)$
		\State Add $(t, t_\text{ref})$ to $Q_\text{ref}$
		
		\State $\dot{L} \leftarrow (L_{t} - L_{t-1})/\delta_{t-1}$
		\If{$\|\dot{L}\| > \varepsilon_\text{max}$}
		\State $\delta_t \leftarrow \min(\delta_{t-1} + \Delta\delta, \delta_\text{max})$
		\ElsIf{$\|\dot{L}\| < \varepsilon_\text{min}$}
		\State $\delta_t \leftarrow \max(\delta_{t-1} - \Delta\delta, \delta_\text{min})$
		\Else
		\State $\delta_t \leftarrow \delta_{t-1}$
		\EndIf
		
		\If{$\exists (t', t) \in Q_\text{ref}$}
		\State Get all pairs $(t', t)$ from $Q_\text{ref}$
		\For{each $(t', t)$ pair}
		\State Get $\boldsymbol{\theta}_{t'-1}$ from $Q_{\theta}$
		\State $\Delta \boldsymbol{\theta}_{t'} \leftarrow \boldsymbol{\theta}_{t} - \boldsymbol{\theta}_{t'-1}$
		\For{each $i \in \mathcal{B}_{t'}$}
		\State $\mathbf{g}_i \leftarrow \nabla_{\boldsymbol{\theta}} L(f(\mathbf{x}_i; \boldsymbol{\theta}_{t'-1}), y_i)$
		\State $\mathbf{u}_i^{t'} \leftarrow \boldsymbol{\theta}_{t} - (\boldsymbol{\theta}_{t'-1} - \eta_{t'} \mathbf{g}_i)$
		\State $v_i^{t'} \leftarrow \frac{\|\Delta \boldsymbol{\theta}_t\| - \|\mathbf{u}_i^{t'}\|}{\|\Delta \boldsymbol{\theta}_t\| + \|\mathbf{u}_i^{t'}\|}$
		\State $v_i^{[0,T]} \leftarrow v_i^{[0,T]} + v_i^{t'}$
		\EndFor
		\State Remove $(t', t)$ from $Q_\text{ref}$
		\EndFor
		\EndIf
		
		\State Remove from $Q_{\theta}$ where step $< t - \delta_\text{max}$
		\State Remove from $Q_\text{ref}$ where $t_\text{ref} < t$
		\EndFor
		\State \Return $\{v_i^{t}\}_{t=1}^T, v_i^{[0,T]}$
	\end{algorithmic}
\end{algorithm}

During each training iteration, LiveVal orchestrates a series of operations to maintain efficient real-time valuation. The system first performs reference point checking to identify when stored parameters become reference points for previous states. When matches occur, it computes data values for the corresponding historical mini-batches and updates the adaptive window size based on observed loss dynamics. Queue maintenance then removes processed states to ensure memory efficiency.

This efficient design brings significant improvements in both memory and computational complexity. The memory requirement is reduced from $O(Td)$ for storing complete training trajectories to $O(\delta_{\text{max}}d + N + T)$, where $d$ is the model dimension and $N$ is the dataset size. The per-step time complexity remains bounded at $O(\delta_{\text{max}} + Bd)$, where $B$ is the batch size, enabling efficient processing even with large models and datasets. Algorithm \ref{alg:liveval_integrated} in Appendix \ref{app:implementation} provides the complete implementation details.

\section{Theoretical Analysis}
\subsection{Fundamental Properties}
We first establish that LiveVal's valuations meaningfully reflect data contributions through two core properties.
\begin{theorem}[Fundamental Properties of LiveVal] \label{theoremfundamental_properties}
	For any data point \( i \) and iteration \( t \), the step-wise data value \( v_i^t \) of LiveVal satisfies:
	
	1) Directional Alignment: If the gradient update from data point \( i \) moves the model parameters closer to the reference point \( \boldsymbol{\theta}_{\text{ref}}^t \), then \( v_i^t \geq 0 \).
	
	2) Value Boundedness: For all data points \( i \) and iterations \( t \), \( v_i^t \in [-1, 1] \).
\end{theorem}
The detailed analysis is presented in Appendix \ref{app:proofs}. 
These properties ensure interpretable valuations, as positive values indicate beneficial updates while the bounded range enables fair comparison across training stages.

\subsection{Stability Analysis}
We further show that LiveVal's valuations maintain stability during training through bounded volatility.
\begin{theorem}[Local Volatility Bound] \label{thmlocal_bound}
	Suppose that for all model parameters $\boldsymbol{\theta}$ and data points $i$, the gradient norms are bounded by $\|\nabla_{\boldsymbol{\theta}} L(f(\mathbf{x}_i; \boldsymbol{\theta}), y_i)\| \leq G$, and the distance to the reference point satisfies $\|\Delta \boldsymbol{\theta}_t\| \geq D_{\min} > 0$. Then, for any data point $i$ at training step $t$, the volatility of the step-wise data valuation $v_i^{t}$ is bounded by:
	\begin{equation}
		\sigma_i^{[t]} \leq \frac{2 \eta_t G}{\|\Delta \boldsymbol{\theta}_t\|},
	\end{equation}
	where $\eta_t$ is the learning rate at step $t$.
\end{theorem}
This stability guarantee shows that valuations remain reliable as training progresses, with larger distances to reference points leading to more stable valuations. Detailed proofs are in Appendix \ref{app:proofs}.

\section{Experiments}
We evaluate LiveVal in three aspects: accuracy and efficiency in identifying valuable samples compared to baselines (Section \ref{seccomp}, and \ref{secedc}), robustness across different data qualities and modalities (Section \ref{seccor} and \ref{seccrxmdl}), and scalability to large-scale models (Section \ref{seclarge}).

\subsection{Experimental Setup}
\paragraph {Datasets and Models} Our evaluation uses datasets spanning different modalities: Adult (tabular), 20 Newsgroups (text), and MNIST (images), tested on DNN, CNN (LeNet-5), and large-scale Inception V3 (24M parameters).

\paragraph {Baselines} We compare against three methods:

\textbf{1) Leave-One-Out (LOO)}: Computes a data point's value by measuring its marginal contribution to model loss:
\begin{equation}
	v_i^{\text{LOO}} = L(\mathcal{D}\setminus\{i\}) - L(\mathcal{D})
\end{equation}
where $L(\mathcal{D})$ is the model's loss on the full dataset and $L(\mathcal{D}\setminus\{i\})$ is the loss after removing point $i$. LOO requires retraining the model for each data point.

\textbf{2) Influence Functions (IF)} \cite{koh2017understanding}: Approximates data value through the model's loss change:
\begin{equation}
	v_i^{\text{IF}} = -\nabla_{\boldsymbol{\theta}} L(\mathbf{z}_{\text{test}}, \boldsymbol{\theta}^*)^\top H_{\boldsymbol{\theta}^*}^{-1} \nabla_{\boldsymbol{\theta}} L(\mathbf{x}_i, \boldsymbol{\theta}^*)
\end{equation}
where $H_{\boldsymbol{\theta}^*}$ is the Hessian matrix at optimal parameters $\boldsymbol{\theta}^*$.

\textbf{3) GradNd} \cite{paul2021deep}: A real-time method that values data points based on their gradient during training:
\begin{equation}
	v_i^{\text{GradNd}} = \mathbb{E}_{\boldsymbol{\theta}_t}[\|\nabla_{\boldsymbol{\theta}} L(\mathbf{x}_i, \boldsymbol{\theta}_t)\|_2]
\end{equation}

Experiments are conducted using PyTorch 2.0 on NVIDIA RTX 2080 Ti GPUs, with five random seeds for statistical significance. Detailed experimental settings are provided in the Appendix \ref{secsetup}.

\subsection{Comparative Analysis Against Baselines} \label{seccomp}
A key objective of data valuation methods is to identify harmful or beneficial samples. This section compares LiveVal's valuation accuracy and resource efficiency against baselines through a label corruption detection task.

\paragraph{Experimental Design}
We select MNIST dataset for this analysis. Specifically, we flip $k$ labels from digit `1' to `7' ($k \in \{10, 20, 30, 40\}$), chosen for their structural similarities to create a challenging detection scenario. Using modified MNIST to train a CNN model, we evaluate data values on 100 samples ($k$ corrupted samples and $(100-k)$ randomly selected clean samples) and check the number of flipped samples among the $k$ lowest-valued samples.

\begin{table}[h]
	\centering
	\caption{Detection Performance on MNIST Label Corruption Task}
	\label{tab:detection}
	\resizebox{\linewidth}{!}{
		\begin{tabular}{ccccc}
			\toprule
			\textbf{\# Flipped} & \textbf{LOO} & \textbf{IF} & \textbf{GradNd} & \textbf{LiveVal} \\
			\midrule
			10 & 8.4 $\pm$ 2.4&4.6 $\pm$ 2.1 &0.2 $\pm$ 0.4& \textbf{6.4 $\pm$ 2.3}\\
			20 & 17.4 $\pm$ 1.7 & 11.8 $\pm$ 4.4 & 0.6 $\pm$ 0.9&\textbf{13.8 $\pm$ 1.1}\\
			30 & 27.0 $\pm$ 1.4 & 18.0 $\pm$ 4.0 & 0.8 $\pm$ 1.3 &\textbf{20.6 $\pm$ 1.7} \\
			40 & 37.2 $\pm$ 2.2 & 24.2 $\pm$ 4.0& 1 $\pm$ 0.7 & \textbf{28.4 $\pm$ 1.5} \\
			\bottomrule
		\end{tabular}
	}
\end{table}

\paragraph{Results and Analysis} 
Table \ref{tab:detection} shows LiveVal's strong detection capability. While LOO achieves the best detection accuracy (84\% at $k=10$) through exhaustive retraining, LiveVal achieves comparable performance (64\%) with significantly lower resource requirements. The performance gap between LiveVal and GradNd, IF demonstrates the importance of time-aware data valuation.

\begin{table}[h]
	\small
	\centering
	\caption{Resource Requirements for Different Methods}
	\label{tab:resources}
	\setlength{\tabcolsep}{2pt}  
	\renewcommand{\arraystretch}{1.2}  
	\begin{tabular}{ccccc}
		\toprule
		\textbf{Metrics} & \textbf{LOO} & \textbf{IF} & \textbf{GradNd} & \textbf{LiveVal} \\
		\midrule
		Computation Time & 3h 34m 29s & 1m 29s & 13m 2s & \textbf{1m 11s} \\
		\makecell{Graphics\\Memory Usage} & 6.817GB & 0.521 GB & 0.522GB & \textbf{0.521GB} \\
		\bottomrule
	\end{tabular}
\end{table}

As shown in Table \ref{tab:resources}, LiveVal significantly reduces resource requirements compared to other methods. It achieves a 180× speedup over LOO (71s vs 3.5h) through seamless integration with SGD training. Memory efficiency is measured using \texttt{nvidia-smi} to track peak GPU usage during evaluation. LiveVal's adaptive reference mechanism enables minimal memory usage (0.521GB vs. LOO's 6.817GB) by discarding outdated parameters, making it particularly suitable for large-scale applications where computational resources are constrained. While IF shows similar speed, it requires model convergence before evaluation and achieves lower accuracy due to its restrictive assumptions. GradNd's poor performance confirms that local gradient information alone is insufficient for accurate data valuation.

\subsection{Early Detection Capability} \label{secedc}
A unique advantage of LiveVal is its ability to perform detection during training. Using the same label corruption experiments described above, we analyze LiveVal's detection performance across training epochs. 

\begin{figure}[h]
	\centering
	\includegraphics[width=0.99\linewidth]{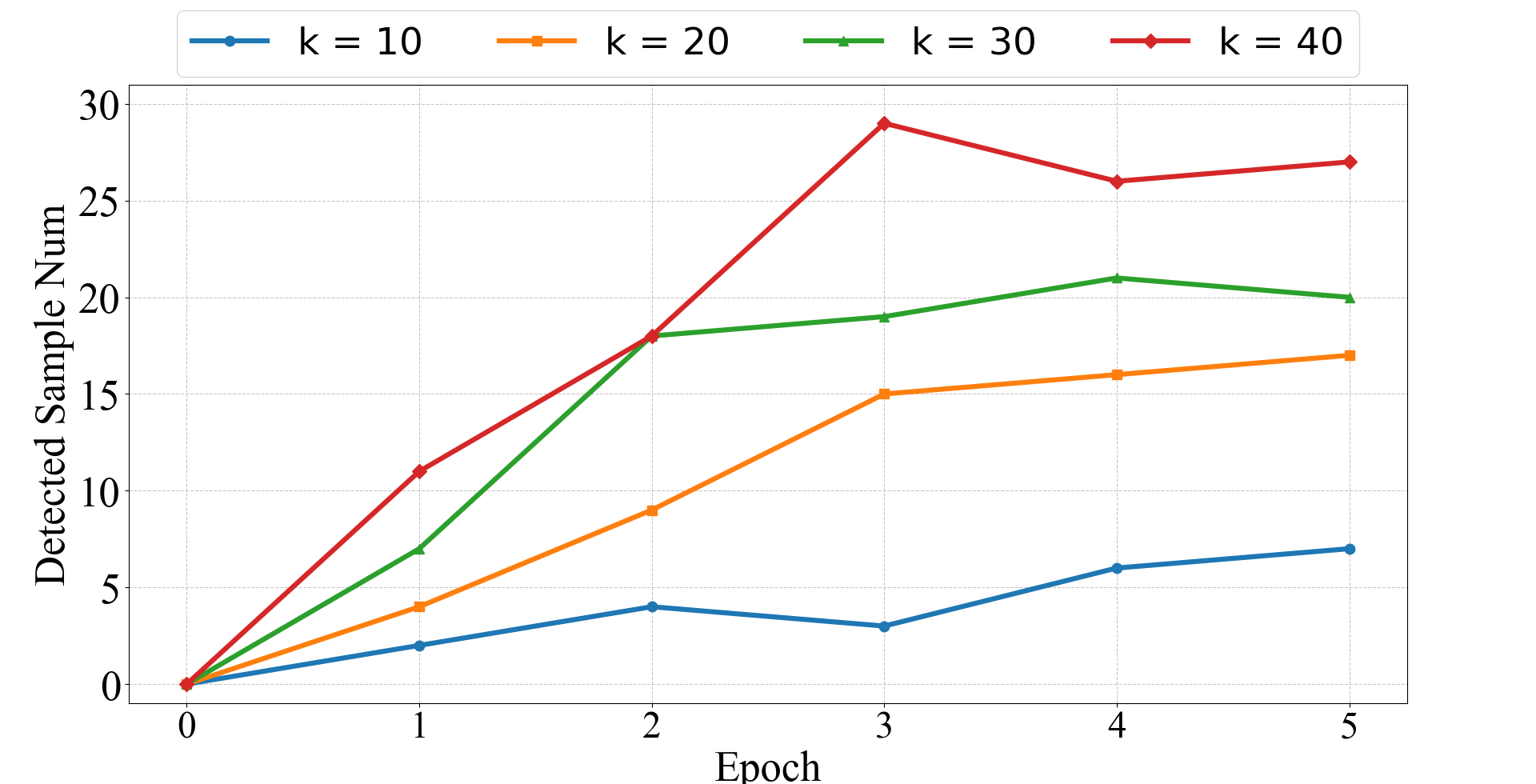}
	\caption{Early Detection Performance of LiveVal. The curves show the number of identified corrupted samples in the first 5 training epochs for different corruption levels ($k=10,20,30,40$). }
	\label{fig:epoch_detection}
\end{figure}

Figure \ref{fig:epoch_detection} demonstrates LiveVal's strong early detection capability. For the most challenging scenario (k=40), LiveVal identifies 27.5\% of corrupted samples within the first epoch and achieves 72.5\% detection rate by epoch 3. This early detection capability, combined with the stable performance across different flipped samples, further validates LiveVal's efficiency and practical utility for real-time data valuation.

\subsection{Robustness Analysis of Feature Corruption} \label{seccor}
Beyond label corruption, we evaluate LiveVal's capability to detect feature-level data quality issues, which present more subtle detection challenges than label errors.

\paragraph{Experimental Design}
This experiment is specifically designed to assess LiveVal's sensitivity to different degrees of data corruption and its robustness across varying scales of corruption. Using the MNIST dataset with LeNet-5 model, we inject Gaussian noise with different standard deviations $\sigma \in \{1.0, 2.0, 5.0\}$ to $k$ randomly selected samples ($k \in \{10, 20, 30, 40\}$). We train the LeNet-5 model on this modified dataset and compute data values using LiveVal. We evaluate detection performance by checking how many corrupted samples appear in the $k$ lowest-valued samples among 100 test samples (all $k$ corrupted plus $(100-k)$ clean samples).

\begin{table}[h]
	\centering
	\caption{Number of Detected Samples Under Feature Corruption}
	\label{tab:corruption_detection}
	\begin{tabular}{clll}
		\toprule
		\textbf{\# Corrupted} & \textbf{$\sigma$ = 1.0} & \textbf{$\sigma$ = 2.0} & \textbf{$\sigma$ = 5.0} \\
		\midrule
		10 & 2.0 ± 1.6 & 2.4 ± 1.1 & 2.4 ± 1.1 \\
		20 & 6.2 ± 4.4 & 9.0 ± 2.6 & 10.2 ± 2.3 \\
		30 & 11.2 ± 6.0 & 13.2 ± 3.1 & 18.0 ± 4.2 \\
		40 & 25.8 ± 2.9 & 28.4 ± 6.2 & 30.0 ± 4.7 \\
		\bottomrule
	\end{tabular}
\end{table}

\paragraph{Results and Analysis}
Table \ref{tab:corruption_detection} reveals two key properties of LiveVal. First, LiveVal shows strong adaptability to different levels of data corruption, with detection rates increasing as noise magnitude grows (e.g., from 11.2 to 18.0 samples at $k=30$ when $\sigma$ increases from 1.0 to 5.0). This sensitivity stems from LiveVal's parameter space analysis: larger noise leads to more significant parameter deviations from the reference trajectory, making corrupted samples more distinguishable through their gradient contributions.

Second, LiveVal performs better with larger-scale corruptions, identifying 75\% of corrupted samples at $k=40$ compared to 24\% at $k=10$ under $\sigma=5.0$. This scale-dependent behavior arises from LiveVal's adaptive reference mechanism: when more corrupted samples are present, their collective impact creates more distinct parameter trajectories, enabling more reliable detection through reference point comparisons. These properties make LiveVal particularly effective for real-world scenarios where data quality issues often manifest as systematic corruptions affecting multiple samples rather than isolated noise.

\subsection{Robustness Analysis of Different Modalities } \label{seccrxmdl}
Different data modalities present distinct challenges for data valuation due to their inherent structural differences. We evaluate LiveVal's generalizability across text and tabular data modalities.
\paragraph{Experimental Design}
We evaluate LiveVal on two additional datasets: 20 Newsgroups (text) and Adult (tabular), using a DNN architecture. We deliberately flip k\% of labels ($k \in \{10,20,30,40\}$) in each dataset. Following our previous evaluation method, for each $k$, we evaluate detection performance on 100 samples ($k$ flipped plus $(100-k)$ randomly selected clean samples) and examine how many flipped samples appear in the $k$ lowest-valued samples. 

\begin{table}[h]
	\small
	\centering
	\caption{Number of Detected Flipped Labels Across Modalities}
	\label{tab:detection_Modalities}
	\setlength{\tabcolsep}{2pt} 
	\renewcommand{\arraystretch}{1.2} 
	\begin{tabular}{ccccc}
		\toprule
		\textbf{\# Flipped} & \textbf{k=10} & \textbf{k=20} & \textbf{k=30} & \textbf{k=40} \\
		\midrule
		20Newsgroups & 3.0 ± 1.0 & 8.8 ± 2.3 & 16.8 ± 3.4 & 22.6 ± 2.4 \\
		Adult & 3.4 ± 2.1 & 6.8 ± 1.9 & 11.6 ± 2.4 & 17.2 ± 3.4 \\
		\bottomrule
	\end{tabular}
\end{table}

\paragraph{Results and Analysis}
Table \ref{tab:detection_Modalities} reveals LiveVal's varying effectiveness across different data modalities. For text data (20 Newsgroups), LiveVal's detection rate improves substantially from 30\% at $k=10$ to 56.5\% at $k=40$. This strong scaling behavior can be attributed to the rich feature interactions in text data: corrupted samples tend to create more distinct parameter trajectories due to the high-dimensional, sparse nature of text representations. 
For tabular data (Adult), LiveVal shows more moderate improvement, from 34\% at $k=10$ to 43\% at $k=40$. Despite these modality-specific variations, LiveVal maintains good detection performance across both domains, demonstrating its potential as a general-purpose data valuation framework.

\subsection{Scaling to Large Models} \label{seclarge}
Modern deep learning often employs large architectures where traditional data valuation methods become computationally impractical. We evaluate LiveVal on Inception V3 (48 layers with 24M parameters) to show its scalability.

\paragraph{Experimental Design}
Following the same label flipping setup from Section \ref{seccomp}, we flip $k$ labels from digit `1' to `7' where $k \in \{10,20,30,40\}$ and train the Inception V3 model using the modified MNIST dataset for one epoch. We evaluate randomly selected 100 samples, including $k$ flipped and $(100-k)$ clean samples, and identify the $k$ lowest-valued samples. 

\begin{table}[h]
	\small
	\centering
	\caption{LiveVal's Flipped Label Detection on Large-scale Model}
	\label{tab:detection_large}
	\setlength{\tabcolsep}{2pt} 
	\begin{tabular}{ccccc}
		\toprule
		\textbf{\# Flipped} & \textbf{k=10} & \textbf{k=20} & \textbf{k=30} & \textbf{k=40} \\
		\midrule
		\textbf{\# Detected } & 3.0 ± 2.2 & 8.6 ± 4.2 & 11.4 ± 5.9 & 22.6 ± 2.4 \\
		\textbf{Detect Rate } & 30\% & 43\% & 38\% & 56\% \\
		\bottomrule
	\end{tabular}
\end{table}

\paragraph{Results and Analysis} 
The experimental results demonstrate LiveVal's effectiveness on large-scale models from multiple aspects. Table \ref{tab:detection_large} shows LiveVal's detection performance on Inception V3 with 24M parameters. With increasing corruption levels, LiveVal maintains stable detection rates, reaching 56\% for $k=40$ even with just one epoch of training. This early detection capability is crucial for identifying harmful samples before they significantly impact the model training.

Beyond detection accuracy, our evaluation also reveals significant computational advantages. While not shown in the table, LiveVal completes the evaluation in $7\,\text{min}\, 32\text{s}$ on a single NVIDIA A40 GPU, a remarkable achievement considering Inception V3's 24M parameters, compared to LOO's runtime of hundreds of hours. Additionally, LiveVal requires only 31.18GB compared to influence functions' 2.3TB for Hessian computation.
These efficiency gains, combined with the detection capability, make LiveVal practical for large-scale architectures where traditional methods become prohibitive.

\section{Related Work}
Data valuation methods aim to quantify the contribution of individual training samples to model performance. Leave-One-Out (LOO) methods~\cite{cook1977detection,sharchilev2018finding} directly measure contributions by training separate models with and without target samples. However, this approach faces prohibitive computational costs for large datasets and inherent instability from training randomness~\cite{sim2022data}.
Game-theoretic approaches like Shapley values~\cite{shapley1953,wang2020principled} extend LOO by measuring marginal contributions across all possible subset combinations. While theoretically sound, these methods require evaluating exponential subset combinations, making them impractical for large-scale learning~\cite{ghorbani2019data,sun2023shapleyfl,kwon2022beta}. Recent work~\cite{wang2023data,li2024robust} further reveals that both LOO and Shapley methods suffer from inconsistent rankings due to retraining randomness. Although approximation techniques~\cite{luo2024fast,zhang2023efficient} reduce computational overhead, they still require extensive subset evaluations.

Local update methods~\cite{pruthi2020estimating,paul2021deep,tan2023data} use training gradients for efficient estimation but struggle with training order bias, as early samples receive disproportionate importance due to varying gradient magnitudes across training stages. While cosine similarity methods~\cite{miao2022privacy,fung2018mitigating} measure gradient alignments with overall training direction, they fail to capture long-term effects. LiveVal overcomes these limitations through its adaptive reference point mechanism, enabling time-aware contribution measurement while maintaining computational efficiency.

Influence functions~\cite{koh2017understanding} provide theoretical insights but rely on assumptions rarely met in modern deep learning: loss convexity, model optimality, and Hessian computations~\cite{basu2021influence,hammoudeh2024training}. Although recent extensions~\cite{koh2019accuracy,basu2020second,schioppa2022scaling} propose approximations for non-convex scenarios, challenges remain for real-time analysis in deep architectures~\cite{basu2021influence}. In contrast, LiveVal operates effectively without convexity or optimality assumptions, making it suitable for typical deep learning scenarios. 

\section{Conclusion}
We presented LiveVal, a time-aware data valuation framework that enables efficient, real-time assessment of training data quality in large-scale machine learning systems. By transforming valuation from loss computation to parameter space analysis with adaptive reference points, LiveVal overcomes key limitations of existing methods: the computational burden of retraining approaches, the temporal bias of gradient methods, and the restrictive assumptions of influence functions. Our theoretical analysis establishes bounded and stable valuations, while experiments demonstrate LiveVal's ability to identify harmful samples early in training, offering significant computational savings for large-scale applications.   

\bibliographystyle{IEEEtran}
\bibliography{references}

%%%%%%%%%%%%%%%%%%%%%%%%%%%%%%%%%%%%%%%%%%%%%%%%%%%%%%%%%%%%%%%%%%%%%%%%%%%%%%%
%%%%%%%%%%%%%%%%%%%%%%%%%%%%%%%%%%%%%%%%%%%%%%%%%%%%%%%%%%%%%%%%%%%%%%%%%%%%%%%
% APPENDIX
%%%%%%%%%%%%%%%%%%%%%%%%%%%%%%%%%%%%%%%%%%%%%%%%%%%%%%%%%%%%%%%%%%%%%%%%%%%%%%%
%%%%%%%%%%%%%%%%%%%%%%%%%%%%%%%%%%%%%%%%%%%%%%%%%%%%%%%%%%%%%%%%%%%%%%%%%%%%%%
%\newpage
\clearpage
\appendix
 
\section{Implementation} \label{app:implementation}
Algorithm \ref{alg:liveval_infer_until_t} presents the basic time-aware data valuation method.

		\begin{algorithm}[H]
			\caption{Basic Time-Aware Data Valuation}
			\label{alg:liveval_infer_until_t}
			\begin{algorithmic}[1]
				\Require{Training sample $(\mathbf{x}_i, y_i)$, Stored information $A$, Total number of training steps $T$}
				\Ensure{Time series of data valuations $\{v_i^{t}\}_{t=1}^T$, Cumulative data valuation $v_i^{[0,T]}$}
				\State{Initialize $v_i^{[0,T]} \leftarrow 0$}
				\State{Initialize array $\{v_i^{t}\}_{t=1}^T \leftarrow \mathbf{0}$} 
				\State{$\boldsymbol{\theta}^* \leftarrow A[T]$}
				\For{$t = 1$ to $T$}
				\State{$(\boldsymbol{\theta}_{t-1}, \eta_t, \mathcal{B}_t) \leftarrow A[t]$}
				\State{$\Delta \boldsymbol{\theta}_t \leftarrow \boldsymbol{\theta}^* - \boldsymbol{\theta}_{t-1}$}
				\If{$i \in \mathcal{B}_t$}
				\State{$\mathbf{u}_i^t \leftarrow \boldsymbol{\theta}^* - (\boldsymbol{\theta}_{t-1} - \eta_t \nabla_{\boldsymbol{\theta}} L(f(\mathbf{x}_i;{\boldsymbol{\theta}_{t-1}}), y_i))$}
				\State{$v_i^{t} \leftarrow (\|\Delta \boldsymbol{\theta}_t\| - \|\mathbf{u}_i^t\|) / \|\Delta \boldsymbol{\theta}_t\|$}
				\State{$v_i^{[0,T]} \leftarrow v_i^{[0,T]} + v_i^{t}$}
				\EndIf
				\EndFor
				\State \Return $\{v_i^{t}\}_{t=1}^T, v_i^{[0,T]}$
			\end{algorithmic}
		\end{algorithm}

		\begin{algorithm}[t]
			\caption{LiveVal: Time-aware Data Valuation}
			\label{alg:liveval_integrated}
			\begin{algorithmic}[1]
				\Require Training dataset $\{(\mathbf{x}_i, y_i)\}_{i=1}^N$, Total steps $T$, Initial window $\delta_0$, Window bounds $\delta_\text{min}$, $\delta_\text{max}$, Adaptation rate $\Delta \delta$, Loss thresholds $\varepsilon_\text{min}$, $\varepsilon_\text{max}$
				\Ensure Step-wise values $\{v_i^{t}\}_{t=1}^T$, Cumulative value $v_i^{[0,T]}$
				
				\State Initialize $v_i^{[0,T]} \leftarrow 0$ for all $i$
				\State Initialize array $\{v_i^{t}\}_{t=1}^T \leftarrow \mathbf{0}$ for all $i$
				\State Initialize window size $\delta_0$
				\State Initialize window size array $\{\delta_t\}_{t=1}^T \leftarrow \{\delta_0\}$
				\State Initialize model queue $Q_{\theta}$ for storing model parameters
				\State Initialize reference queue $Q_\text{ref}$ for storing $(t_\text{eval}, t_\text{ref})$ pairs
				\State Initialize $\boldsymbol{\theta}_0$
				
				\For{$t = 1$ to $T$}
				\State Sample mini-batch $\mathcal{B}_{t}$
				\State $\mathbf{g}_{t} \leftarrow \frac{1}{\|\mathcal{B}_{t}\|} \sum_{i \in \mathcal{B}_{t}} \nabla_{\boldsymbol{\theta}} L(f(\mathbf{x}_i;\boldsymbol{\theta}_{t-1}), y_i)$
				\State $\boldsymbol{\theta}_{t} \leftarrow \boldsymbol{\theta}_{t-1} - \eta_{t} \mathbf{g}_{t}$
				\State Store $(t, \boldsymbol{\theta}_{t}, \mathcal{B}_{t}, \eta_{t}, L_{t})$ in $Q_{\theta}$
				\State $t_\text{ref} \leftarrow \min(t-1 + \delta_{t-1}, T)$
				\State Add $(t, t_\text{ref})$ to $Q_\text{ref}$
				
				\State $\dot{L} \leftarrow (L_{t} - L_{t-1})/\delta_{t-1}$
				\If{$\|\dot{L}\| > \varepsilon_\text{max}$}
				\State $\delta_t \leftarrow \min(\delta_{t-1} + \Delta\delta, \delta_\text{max})$
				\ElsIf{$\|\dot{L}\| < \varepsilon_\text{min}$}
				\State $\delta_t \leftarrow \max(\delta_{t-1} - \Delta\delta, \delta_\text{min})$
				\Else
				\State $\delta_t \leftarrow \delta_{t-1}$
				\EndIf
				
				\If{$\exists (t', t) \in Q_\text{ref}$}
				\State Get all pairs $(t', t)$ from $Q_\text{ref}$
				\For{each $(t', t)$ pair}
				\State Get $\boldsymbol{\theta}_{t'-1}$ from $Q_{\theta}$
				\State $\Delta \boldsymbol{\theta}_{t'} \leftarrow \boldsymbol{\theta}_{t} - \boldsymbol{\theta}_{t'-1}$
				\For{each $i \in \mathcal{B}_{t'}$}
				\State $\mathbf{g}_i \leftarrow \nabla_{\boldsymbol{\theta}} L(f(\mathbf{x}_i; \boldsymbol{\theta}_{t'-1}), y_i)$
				\State $\mathbf{u}_i^{t'} \leftarrow \boldsymbol{\theta}_{t} - (\boldsymbol{\theta}_{t'-1} - \eta_{t'} \mathbf{g}_i)$
				\State $v_i^{t'} \leftarrow \frac{\|\Delta \boldsymbol{\theta}_t\| - \|\mathbf{u}_i^{t'}\|}{\|\Delta \boldsymbol{\theta}_t\| + \|\mathbf{u}_i^{t'}\|}$
				\State $v_i^{[0,T]} \leftarrow v_i^{[0,T]} + v_i^{t'}$
				\EndFor
				\State Remove $(t', t)$ from $Q_\text{ref}$
				\EndFor
				\EndIf
				
				\State Remove from $Q_{\theta}$ where step $< t - \delta_\text{max}$
				\State Remove from $Q_\text{ref}$ where $t_\text{ref} < t$
				\EndFor
				\State \Return $\{v_i^{t}\}_{t=1}^T, v_i^{[0,T]}$
			\end{algorithmic}
		\end{algorithm}

Algorithm \ref{alg:liveval_integrated} presents LiveVal's framework that integrates SGD training steps with real-time data value computation through a dual-queue architecture. For each SGD training step $t$, the algorithm maintains necessary information for future data value computation and performs value computation when conditions are met.

The dual-queue system uses two synchronized queues: a Model Queue $Q_{\theta}$ and a Reference Queue $Q_{\text{ref}}$. The Model Queue $Q_{\theta}$ serves as a dynamic repository storing model parameters as (SGD\_step, parameter) pairs $(t, \boldsymbol{\theta}_t)$ from recent training steps. This queue maintains a sliding window of parameters needed for future value computation, keeping only parameters within the interval $[t-\delta_{\text{max}}, t]$ where $t$ is the current SGD training step. Given a model with $d$ parameters, the memory complexity of $Q_{\theta}$ is $O(\delta_{\text{max}}d)$, as it stores at most $\delta_{\text{max}}$ model states. 

The Reference Queue $Q_{\text{ref}}$ tracks which historical SGD steps require value computation and their corresponding reference points through $(t_{\text{eval}}, t_{\text{ref}})$ pairs, which are $(t, t_{\text{SGD}})$ in Algorithm \ref{alg:liveval_integrated}. For each SGD step $t_{\text{eval}}$, a future step $t_{\text{ref}} = \min(t_{\text{eval}}-1 + \delta_{t_{\text{eval}}-1}, T)$ is its reference point, where $\delta_{t_{\text{eval}}-1}$ is the adaptive window size at step $t_{\text{eval}}-1$. The queue stores at most $\delta_{\text{max}}$ pairs of indices, resulting in a memory complexity of $O(\delta_{\text{max}})$. The value computation for data points in mini-batch $\mathcal{B}_{t_{\text{eval}}}$ is triggered when the current SGD training reaches step $t_{\text{ref}}$.

The value computation process consists of three stages that occur during SGD training. First, at each current SGD step $t$, the system performs a Reference Point Check by examining if $t$ matches any reference point $t_{\text{ref}}$ stored in $Q_{\text{ref}}$. This operation has time complexity $O(\delta_{\text{max}})$. When matches are found, the system proceeds to the Value Computation, where for each corresponding evaluation step $t_{\text{eval}}$, it computes the value $v_i^{t_{\text{eval}}}$ for all samples $i$ in mini-batch $\mathcal{B}_{t_{\text{eval}}}$ and updates their cumulative values $v_i^{[0, T]}$. Additionally, the system updates the adaptive window size $\delta_{t_{\text{eval}}}$ based on the loss change rate between the evaluation step and the current reference point. With batch size $B$, this stage has time complexity $O(Bd)$ per reference point. Finally, in the Queue Maintenance, the system removes the processed $(t_{\text{eval}}, t_{\text{ref}})$ pairs from $Q_{\text{ref}}$ and cleans up model parameters in $Q_{\theta}$ that are no longer needed for future computations. This operation has time complexity $O(\delta_{\text{max}})$.

The total memory complexity of LiveVal is $O(\delta_{\text{max}}d + N + T)$, where $d$ is the model dimension, $N$ is the dataset size for storing cumulative values, and $T$ is for storing the window size array. This is significantly more efficient than storing the entire training trajectory, requiring $O(Td)$ memory for $T$ training steps. The time complexity per step is $O(\delta_{\text{max}} + Bd)$ in the worst case when value computation is triggered.

This efficient design enables LiveVal to achieve real-time data valuation while maintaining bounded memory overhead, making it particularly suitable for large-scale machine learning applications. The memory requirements scale linearly with the window size $\delta_{\text{max}}$ rather than the total number of training steps, allowing for practical deployment even in resource-constrained environments.
\section{Theoretical Analysis of LiveVal} \label{app:proofs}
This section establishes the theoretical foundations and guarantees of LiveVal. We first formalize the mathematical framework and necessary assumptions. We then prove key properties that ensure reliable data valuation and analyze stability under stochastic training conditions. Our analysis shows that LiveVal provides reliable, bounded data valuations while maintaining robustness to training dynamics.

\subsection{Theoretical Framework}
To facilitate our analysis, we make the following standard assumptions:

\begin{assumption}[Smoothness]
	\label{assumption:smoothness}
	The loss function \( L(\boldsymbol{\theta}; \mathbf{x}, y) \) is \(\beta\)-smooth with respect to \( \boldsymbol{\theta} \); that is, for all \( \boldsymbol{\theta}, \boldsymbol{\theta}' \in \Theta \) and for all \( (\mathbf{x}, y) \in \mathcal{X} \times \mathcal{Y} \):
	\begin{equation}
		\| \nabla_{\boldsymbol{\theta}} L(\boldsymbol{\theta}; \mathbf{x}, y) - \nabla_{\boldsymbol{\theta}} L(\boldsymbol{\theta}'; \mathbf{x}, y) \| \leq \beta \| \boldsymbol{\theta} - \boldsymbol{\theta}' \|.
	\end{equation}
\end{assumption}

\begin{assumption}[Bounded Gradients]
	\label{assumption:bounded_gradients}
	The gradients of the loss function are uniformly bounded; that is, there exists a constant \( G > 0 \) such that for all \( \boldsymbol{\theta} \in \Theta \) and \( (\mathbf{x}, y) \in \mathcal{X} \times \mathcal{Y} \):
	\begin{equation}
		\| \nabla_{\boldsymbol{\theta}} L(\boldsymbol{\theta}; \mathbf{x}, y) \| \leq G.
	\end{equation}
\end{assumption}

\subsection{Fundamental Properties of LiveVal} \label{secprop}
We establish two fundamental properties that guarantee LiveVal's reliability and interpretability for data valuation. These properties ensure that our method provides meaningful, bounded values that accurately reflect each data point's contribution to model training.

\begin{theorem}[Fundamental Properties of LiveVal]
	\label{theorem:fundamental_properties}
	For any data point \( i \) and iteration \( t \), the step-wise data value \( v_i^t \) of LiveVal satisfies:
	\begin{enumerate}
		\item Directional Alignment: If the gradient update from data point \( i \) moves the model parameters closer to the reference point \( \boldsymbol{\theta}_{\text{ref}}^t \), then \( v_i^t \geq 0 \).
		\item Value Boundedness: For all data points \( i \) and iterations \( t \), \( v_i^t \in [-1, 1] \).
	\end{enumerate}
\end{theorem}

\begin{proof}
	Recall that in LiveVal, the step-wise data value \( v_i^t \) is defined as:
	\begin{equation}
		v_i^t = \frac{\| \Delta \boldsymbol{\theta}_t \| - \| \mathbf{u}_i^t \|}{\| \Delta \boldsymbol{\theta}_t \| + \| \mathbf{u}_i^t \|},
	\end{equation}
	where
	\begin{align*}
		&\Delta \boldsymbol{\theta}_t = \boldsymbol{\theta}_{\text{ref}}^t - \boldsymbol{\theta}_{t-1}, \\
		&\mathbf{u}_i^t = \boldsymbol{\theta}_{\text{ref}}^t - \left( \boldsymbol{\theta}_{t-1} - \eta_t \nabla_{\boldsymbol{\theta}} L(\boldsymbol{\theta}_{t-1}; \mathbf{x}_i, y_i) \right).
	\end{align*}
	
	1) Directional Alignment:
	
	Suppose that the gradient update from data point \( i \) moves the model parameters closer to the reference point \( \boldsymbol{\theta}_{\text{ref}}^t \). This means:
	\begin{equation}
		\label{eq:directional_alignmentcondition}
		\| \boldsymbol{\theta}_{\text{ref}}^t - \left( \boldsymbol{\theta}_{t-1} - \eta_t \nabla_{\boldsymbol{\theta}} L(\boldsymbol{\theta}_{t-1}; \mathbf{x}_i, y_i) \right) \| \leq \| \boldsymbol{\theta}_{\text{ref}}^t- \boldsymbol{\theta}_{t-1} \|.
	\end{equation}
	Using our definitions, this is equivalent to:
	\begin{equation}
		\|\mathbf{u}_i^t\| \leq \|\Delta \boldsymbol{\theta}_t\|.
	\end{equation}
	Therefore, the numerator of \( v_i^t \) satisfies:
	\begin{equation}
		\| \Delta \boldsymbol{\theta}_t \| - \| \mathbf{u}_i^t \| \geq 0.
	\end{equation}
	Since both \( \| \Delta \boldsymbol{\theta}_t \| \) and \( \| \mathbf{u}_i^t \| \) are non-negative, the denominator \( \| \Delta \boldsymbol{\theta}_t \| + \| \mathbf{u}_i^t \| > 0 \). Thus:
	\begin{equation}
		v_i^t = \frac{\| \Delta \boldsymbol{\theta}_t \| - \| \mathbf{u}_i^t \|}{\| \Delta \boldsymbol{\theta}_t \| + \| \mathbf{u}_i^t \|} \geq 0.
	\end{equation}
	
	2) Value Boundedness:
	
	We need to show that \( v_i^t \leq 1 \) and \( v_i^t \geq -1 \) for all \( i \) and \( t \).
	
	First, since \( \| \mathbf{u}_i^t \| \geq 0 \), we have:
	\begin{equation}
		\| \Delta \boldsymbol{\theta}_t \| - \| \mathbf{u}_i^t \| \leq \| \Delta \boldsymbol{\theta}_t \| + \| \mathbf{u}_i^t \|.
	\end{equation}
	Therefore:
	\begin{equation} \label{up}
		v_i^t = \frac{\| \Delta \boldsymbol{\theta}_t \| - \| \mathbf{u}_i^t \|}{\| \Delta \boldsymbol{\theta}_t \| + \| \mathbf{u}_i^t \|} \leq 1.
	\end{equation}
	Similarly, we have:
	\begin{equation}
		\| \mathbf{u}_i^t \| - \| \Delta \boldsymbol{\theta}_t \| \leq \| \Delta \boldsymbol{\theta}_t \| + \| \mathbf{u}_i^t \|.
	\end{equation}
	Multiplying both sides by \( -1 \), we have:
	\begin{equation}
		\| \Delta \boldsymbol{\theta}_t \| - \| \mathbf{u}_i^t \| \geq - ( \| \Delta \boldsymbol{\theta}_t \| + \| \mathbf{u}_i^t \| ).
	\end{equation}
	Therefore:
	\begin{equation}\label{lower}
		v_i^t = \frac{\| \Delta \boldsymbol{\theta}_t \| - \| \mathbf{u}_i^t \|}{\| \Delta \boldsymbol{\theta}_t \| + \| \mathbf{u}_i^t \|} \geq -1.
	\end{equation}
	Combining both Ineq. (\ref{up}) and Ineq. (\ref{lower}) bounds:
	\begin{equation}
		-1 \leq v_i^t \leq 1.
	\end{equation}
\end{proof}

The Directional Alignment property ensures LiveVal correctly identifies helpful training samples. If a data point's gradient update moves the model closer to the reference point, it receives a non-negative value, reflecting a positive contribution. The Value Boundedness property prevents any single data point from having excessive influence, enhancing the robustness of LiveVal against outliers. Together, they make LiveVal's valuations both meaningful and comparable across different training stages.

\subsection{Stability of LiveVal}
Training deep learning models involves randomness from mini-batch sampling, parameter initialization, and optimization dynamics. This randomness could potentially lead to unstable or unreliable data valuations. In this section, we prove that LiveVal is a stable valuation method that provides consistent measurements despite these random factors, ensuring that important data points are reliably identified. 

We aim to show that LiveVal's data valuations remain stable despite training randomness. We do this by 1) analyzing how much a single data point's gradient can change the valuation, 2) using this to bound the maximum possible deviation in valuations, and 3) showing that this bound shrinks as the learning rate decreases.

\begin{definition}[Data Valuation Volatility]
	\label{def:volatility}
	The volatility of the data valuation $v_i^{t}$ for data point $i$ at training step $t$ is defined as the standard deviation of $v_i^{t}$ due to training randomness:
	\begin{equation}
		\sigma_i^{[t]} = \sqrt{\mathbb{E}\left[ \left( v_i^{t} - \mathbb{E}[v_i^{t}] \right)^2 \right]},
	\end{equation}
	where the expectation is taken over all sources of randomness in the training process, including mini-batch sampling and parameter initialization.
\end{definition}
High volatility in valuations could lead to unreliable measurements of data contribution. 

\begin{theorem}[Local Volatility Bound]
	\label{thm:local_bound}
	Suppose that for all model parameters $\boldsymbol{\theta}$ and data points $i$, the gradient norms are bounded by $\|\nabla_{\boldsymbol{\theta}} L(f(\mathbf{x}_i; \boldsymbol{\theta}), y_i)\| \leq G$, and the distance to the reference point satisfies $\|\Delta \boldsymbol{\theta}_t\| \geq D_{\min} > 0$. Then, for any data point $i$ at training step $t$, the volatility of the step-wise data valuation $v_i^{t}$ is bounded by:
	\begin{equation}
		\sigma_i^{[t]} \leq \frac{2 \eta_t G}{\|\Delta \boldsymbol{\theta}_t\|},
	\end{equation}
	where $\eta_t$ is the learning rate at step $t$.
\end{theorem}

\begin{proof}
	We break the proof into three key steps.
	
	Step 1: Bound how much a single gradient update can change the parameter distance.
	
	Recall that in LiveVal, the step-wise data valuation for data point $i$ at step $t$ is computed as:
	\begin{equation}
		v_i^{t} = \frac{\|\Delta \boldsymbol{\theta}_t\| - \|\mathbf{u}_i^{t}\|}{\|\Delta \boldsymbol{\theta}_t\| + \|\mathbf{u}_i^{t}\|},
	\end{equation}
	where 
	\begin{align}
		&\Delta \boldsymbol{\theta}_t = \boldsymbol{\theta}_{\text{ref}}^{t} - \boldsymbol{\theta}_{t-1}, \\
		&\mathbf{u}_i^{t} = \Delta \boldsymbol{\theta}_t + \eta_t \nabla_{\boldsymbol{\theta}} L(f(\mathbf{x}_i; \boldsymbol{\theta}_{t-1}), y_i).
	\end{align}
	The randomness in $v_i^{t}$ arises from the stochastic nature of $\Delta \boldsymbol{\theta}_t$ and $\nabla_{\boldsymbol{\theta}} L(f(\mathbf{x}_i; \boldsymbol{\theta}_{t-1}), y_i)$ due to mini-batch sampling and parameter initialization. To bound $\sigma_i^{[t]}$, we first bound the absolute deviation of $v_i^{t}$ from its expected value.
	
	Using the reverse triangle inequality, we have:
	\begin{equation}
		\left\| \|\mathbf{u}_i^{t}\| - \|\Delta \boldsymbol{\theta}_t\| \right\| \leq \|\mathbf{u}_i^{t} - \Delta \boldsymbol{\theta}_t\| = \eta_t \left\| \nabla_{\boldsymbol{\theta}} L(f(\mathbf{x}_i; \boldsymbol{\theta}_{t-1}), y_i) \right\|.
	\end{equation}
	Since $\|\nabla_{\boldsymbol{\theta}} L(f(\mathbf{x}_i; \boldsymbol{\theta}_{t-1}), y_i)\| \leq G$ by assumption, it follows that:
	\begin{equation}
		\left\| \|\mathbf{u}_i^{t}\| - \|\Delta \boldsymbol{\theta}_t\| \right\| \leq \eta_t G.
	\end{equation}
	Step 2: Show how this parameter change affects the valuation score
	
	We use the triangle inequality:
	\begin{align} \label{eq11}
		\|\mathbf{u}_i^{t} \|&=\| \Delta \boldsymbol{\theta}_t + \eta_t \nabla_{\boldsymbol{\theta}} L(f(\mathbf{x}_i; \boldsymbol{\theta}_{t-1}), y_i)\|\\
		&\geq \| \Delta \boldsymbol{\theta}_t \|- \eta_t \|\nabla_{\boldsymbol{\theta}} L(f(\mathbf{x}_i; \boldsymbol{\theta}_{t-1}), y_i)\|\\
		&\geq \| \Delta \boldsymbol{\theta}_t \|- \eta_t G 
	\end{align} 
	The denominator of $v_i^t$ can be bounded using the triangle inequality:
	\begin{equation} \label{eqde}
		\|\Delta \boldsymbol{\theta}_t\| + \|\mathbf{u}_i^{t}\| \geq \|\Delta \boldsymbol{\theta}_t\| + \left( \|\Delta \boldsymbol{\theta}_t\| - \eta_t G \right) = 2 \|\Delta \boldsymbol{\theta}_t\| - \eta_t G.
	\end{equation}
	
	Assuming that $\eta_t G \leq \|\Delta \boldsymbol{\theta}_t\|$, which holds when the learning rate is sufficiently small or $\|\Delta \boldsymbol{\theta}_t\|$ is sufficiently large, we have:
	\begin{equation} \label{eqnormi}
		2 \|\Delta \boldsymbol{\theta}_t\| - \eta_t G \geq \|\Delta \boldsymbol{\theta}_t\|.
	\end{equation}
	
	By combining Ineq. (\ref{eq11}), Ineq. (\ref{eqnormi}) and Ineq. (\ref{eqnormi}), the absolute value of $v_i^{t}$ can be bounded as:
	\begin{equation}
		\left\| v_i^{t} \right\| = \frac{\left\| \|\Delta \boldsymbol{\theta}_t\| - \|\mathbf{u}_i^{t}\| \right\|}{\|\Delta \boldsymbol{\theta}_t\| + \|\mathbf{u}_i^{t}\|} \leq \frac{\eta_t G}{2 \|\Delta \boldsymbol{\theta}_t\| - \eta_t G} \leq \frac{\eta_t G}{\|\Delta \boldsymbol{\theta}_t\|}.
	\end{equation}
	Step 3: Bound the overall volatility.
	
	Because
	\begin{align}
		\operatorname{Var}(v_i^{t}) &= \mathbb{E}\left[ \left( v_i^{t} - \mathbb{E}[v_i^{t}] \right)^2 \right] \\
		&= \mathbb{E}\left[ (v_i^{t})^2 \right] - \left( \mathbb{E}[v_i^{t}] \right)^2 \leq \mathbb{E}\left[ (v_i^{t})^2 \right].
	\end{align}
	The variance of $v_i^{t}$ is then bounded by:
	\begin{equation}
		\sigma_i^{[t]} = \sqrt{\operatorname{Var}(v_i^{t})} \leq \sqrt{\mathbb{E}\left[ (v_i^{t})^2 \right]} \leq \frac{\eta_t G}{\|\Delta \boldsymbol{\theta}_t\|}.
	\end{equation}
\end{proof}

To make the bound more explicit, we consider the worst-case scenario where $\|\Delta \boldsymbol{\theta}_t\|$ attains its minimum value $D_{\min}$. Thus,
\begin{equation}
	\sigma_i^{[t]} \leq \frac{\eta_t G}{D_{\min}}.
\end{equation}

Since both $G$ and $D_{\min}$ are constants independent of the training step $t$, the volatility $\sigma_i^{[t]}$ is directly proportional to the learning rate $\eta_t$. As $\eta_t$ decreases during training, the volatility of the data valuation $v_i^{t}$ correspondingly decreases. 

This bound indicates that the stability of the data valuations improves as the learning rate decays, which is a common practice in training deep learning models. Therefore, LiveVal provides increasingly reliable data valuations as training progresses.

\section{Experiments } \label{secsetup}
\subsection{Experimental Setup} 
\subsubsection{Datasets} 
We evaluate LiveVal on datasets spanning different modalities and scales:
\begin{itemize}
	\item Adult\cite{Dua2019}: A tabular dataset for binary income prediction containing 48,842 samples with 14 features. This dataset presents challenges in handling mixed categorical and numerical features.
	
	\item 20 Newsgroups \cite{Lang95}: A text classification dataset with 18,846 documents across 20 categories. We preprocess the text using TF-IDF vectorization with 10,000 features, capturing high-dimensional sparse representations typical in NLP tasks.
	
	\item MNIST \cite{lecun-mnisthandwrittendigit-2010}: A handwritten digit classification dataset with 60,000 training and 10,000 testing images (28×28 pixels).
	
\end{itemize}

\subsubsection{Model Architectures} 
We use three architectures of increasing complexity:
\begin{itemize}
	\item Deep Neural Network (DNN): A fully connected network with two hidden layers, ReLU activation, batch normalization, and dropout. 
	
	\item Convolutional Neural Network (CNN): LeNet-5 architecture modified with batch normalization and modern activation functions (ReLU), applied to MNIST.
	
	\item Inception V3: A large-scale deep CNN with 48 layers (about 24 million parameters) featuring inception modules and auxiliary classifiers. This architecture represents modern deep networks and is used to evaluate LiveVal's scalability to complex models.
\end{itemize}

\subsubsection{Baseline Methods} 
We compare against three representative methods.
\begin{itemize}
	\item \textbf{Leave-One-Out (LOO)}: Computes a data point's value by measuring its marginal contribution to model performance:
	\begin{equation}
		v_i^{\text{LOO}} = L(\mathcal{D}\setminus\{i\}) - L(\mathcal{D})
	\end{equation}
	where $L(\mathcal{D})$ is the model's loss on the full dataset and $L(\mathcal{D}\setminus\{i\})$ is the loss after removing point $i$. This requires retraining the model for each data point.
	
	\item \textbf{Influence Functions (IF)} \cite{koh2017understanding}: Approximates data value through the model's loss change when upweighting a training point:
	\begin{equation}
		v_i^{\text{IF}} = -\nabla_{\boldsymbol{\theta}} L(\mathbf{z}_{\text{test}}, \boldsymbol{\theta}^*)^\top H_{\boldsymbol{\theta}^*}^{-1} \nabla_{\boldsymbol{\theta}} L(\mathbf{x}_i, \boldsymbol{\theta}^*)
	\end{equation}
	where $H_{\boldsymbol{\theta}^*}$ is the Hessian matrix at optimal parameters $\boldsymbol{\theta}^*$.
	
	\item \textbf{GradNd} \cite{paul2021deep}: A local update method that values data points based on their gradient magnitude during training:
	\begin{equation}
		v_i^{\text{GradNd}} = \mathbb{E}_{\boldsymbol{\theta}_t}[\|\nabla_{\boldsymbol{\theta}} L(\mathbf{x}_i, \boldsymbol{\theta}_t)\|_2]
	\end{equation}
\end{itemize}

\subsubsection{Evaluation Metrics} 
We evaluate methods through three key dimensions:
\begin{itemize}
	\item \textbf{Valuation Quality.} We measure data valuation accuracy through flipped label detection and feature corruption detection. For flipped label detection, deliberately corrupt $k$ of training samples by randomly changing their labels to incorrect classes. For feature corruption detection, we contaminate features with different Gaussian noise and evaluate detection performance. A good data valuation method should assign lower values to these corrupted samples, enabling their identification.
	\item \textbf{Computational Efficiency.} We measure the wall-clock time required for completing the entire data valuation process.
	\item \textbf{Storage Requirements.} We monitor peak GPU memory consumption during the evaluation process using \texttt{nvidia-smi}, measuring the maximum memory required.
\end{itemize}

\subsubsection{Implementation Details} 
Our implementation uses PyTorch 2.0 with mixed-precision training. Experiments run on three NVIDIA GeForce RTX 2080 Ti. To ensure reproducible performance measurements, we conduct all experiments in a controlled environment where each GPU is exclusively allocated for training. We monitor GPU resource utilization through NVIDIA System Management Interface (\texttt{nvidia-smi}) with 100ms sampling intervals, tracking memory consumption.
All experiments are repeated 5 times with different random seeds to ensure statistical significance. 

\end{document}